\PassOptionsToPackage{table}{xcolor}
\documentclass[lettersize,journal]{IEEEtran}
\usepackage{amsmath,amsfonts}
\usepackage{algorithmic}
\usepackage{algorithm}
\usepackage{array}
\usepackage[caption=false,font=normalsize,labelfont=sf,textfont=sf]{subfig}
\usepackage{textcomp}
\usepackage{stfloats}
\usepackage{url}
\usepackage{verbatim}
\usepackage{graphicx}
\usepackage{cite}
\usepackage{pgfplots}
\pgfplotsset{compat=1.18} 
\usepackage{orcidlink}

\usepackage{multirow}
\usepackage{algorithmic}
\usepackage{algorithm}
\usepackage{makecell}
\usepackage{pgfplots}
\usepackage{bbding} 
\usepackage{placeins}

\usepackage[table]{xcolor}

\begin{document}

\title{DS$^2$Net: Detail-Semantic Deep Supervision Network for Medical Image Segmentation}

\author{Zhaohong Huang, Yuxin Zhang, Taojian Zhou, Guorong Cai and Rongrong Ji
\thanks{Zhaohong Huang, Yuxin Zhang, Rongrong Ji are with the Key Laboratory of Multimedia Trusted Perception and Efficient Computing, Ministry of Education of China,
Xiamen University, Xiamen 361005, China (e-mail: zhaohonghuang@stu.xmu.edu.cn; yuxinzhang@stu.xmu.edu.cn; rrji@xmu.edu.cn).}
\thanks{Taojian Zhou and Guorong Cai are with the Computer Engineering College, Jimei
University, Xiamen 361024, China (e-mail: peacher@jmu.edu.cn; guorongcai@jmu.edu.cn).}
}



\maketitle

\begin{abstract}
Deep Supervision Networks exhibit significant efficacy for the medical imaging community. Nevertheless, existing work merely supervises either the coarse-grained semantic features or fine-grained detailed features in isolation, which compromises the fact that these two types of features hold vital relationships in medical image analysis. We advocate the powers of complementary feature supervision for medical image segmentation, by proposing a Detail-Semantic Deep Supervision Network (DS$^2$Net). DS$^2$Net navigates both low-level detailed and high-level semantic feature supervision through Detail Enhance Module (DEM) and Semantic Enhance Module (SEM). DEM and SEM respectively harness low-level and high-level feature maps to create detail and semantic masks for enhancing feature supervision. This is a novel shift from single-view deep supervision to multi-view deep supervision. DS$^2$Net is also equipped with a novel uncertainty-based supervision loss that adaptively assigns the supervision strength of features within distinct scales based on their uncertainty, thus circumventing the sub-optimal heuristic design that typifies previous works. Through extensive experiments on six benchmarks captured under either colonoscopy, ultrasound and microscope, we demonstrate that DS$^2$Net consistently outperforms state-of-the-art methods for medical image analysis. Our code is available at~\href{https://github.com/hzhxmu/DS-2Net}{https://github.com/hzhxmu/DS-2Net}.
\end{abstract}

\begin{IEEEkeywords}
Medical image segmentation, Deep supervision, Detail and Semantic.
\end{IEEEkeywords}

\section{Introduction}
\IEEEPARstart{M}{edical} image analysis is of paramount importance in contemporary medicine, powering many applications such as treatment planning\cite{duan2022evaluating}, monitoring disease progression\cite{lian2018hierarchical}, and prognosis evaluation\cite{mukherjee2020shallow}.
Particularly, medical image segmentation emerges as a fundamental task that endeavors to implement pixel-level classification within images,~\emph{e.g.}, distinguishing between normal and pathological regions.
However, the segmentation of medical images has proven to be challenging, mostly due to the high complexity of fine-grained features within intricate pathological structures and the variability in coarse-grained semantic information arising from different medical imaging modalities or noise.
To present, the majority of established medical image segmentation models are predicated on the basis of Deep Supervision Network (DSN)\cite{lee2015deeply}, which employs auxiliary outputs at various layers to calculate individual losses in the pursuit of strengthening feature guidance and impacting gradient transformations at multiple network depths, thereby being advisable for the segmentation of medical images. 
According to the supervision mode in DSN, existing studies can be empirically categorized into two principal groups as elucidated below.

\begin{figure}
    \centering
    \includegraphics[width=1\linewidth]{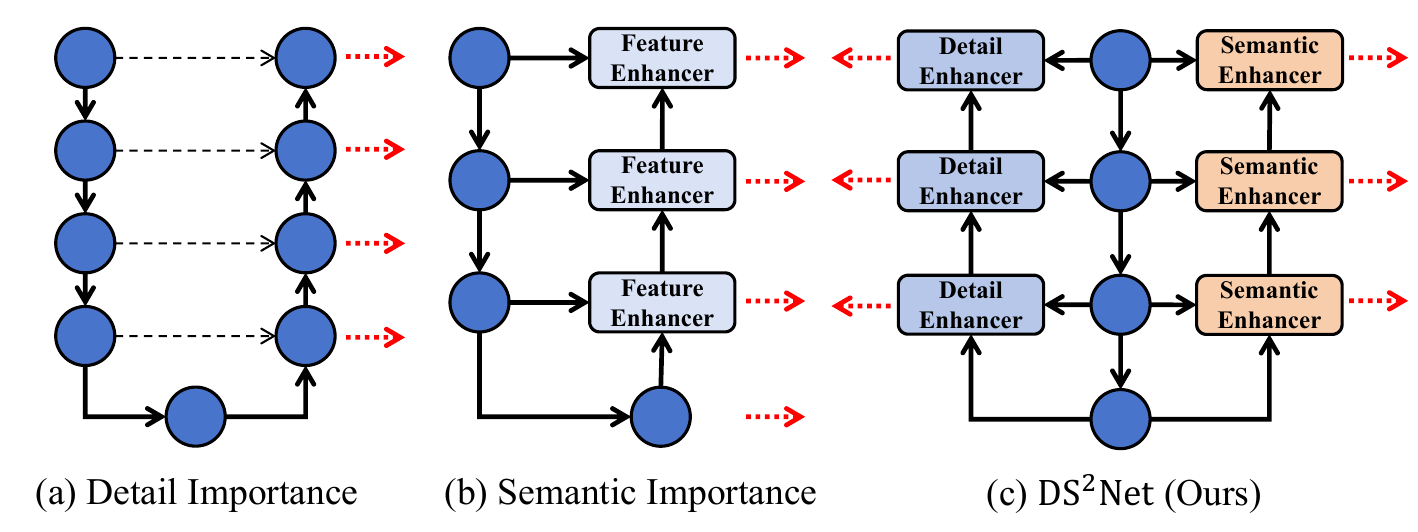}
    \caption{Comparison between different schemes of deep supervision learning. The red dash lines denote the deep supervision and the gray dash lines label the skip connection.}
    \label{fig:motivation}
\end{figure}

\textbf{\textit{Detail Importance:}} As illustrated in Figure\,\ref{fig:motivation}(a), the supervision stems from the output of each stage in accordance with the U-shaped structure\cite{ronneberger2015u}, thereby accentuating the acquisition of fine-grained feature details.
Typical examples include the multiple subnetworks in UNet++\cite{zhou2018unet++}, multi-level shallow feature aggregation in UNet3+\cite{huang2020unet} and the multi-scale input layer in M-Net\cite{fu2018joint}.
While these approaches can effectively discern fine-grained pathological features by examining the shallow features at different stages, the high-level holistic semantic attributes tend to be overlooked.
\textbf{\textit{Semantic Importance:}} This group shifts the focus of deep supervision toward the understanding of high-level pathological semantic information\cite{fan2020pranet,kim2021uacanet}. They typically employ global maps for progressive decoding on the basis of intermediate coarse-grained feature maps to obtain hierarchical signals, as illustrated in Figure\,\ref{fig:motivation}(b). 
For instance, UACANet\cite{kim2021uacanet} considers uncertainty regions of saliency maps, 
while CaraNet\cite{lou2022caranet} employs context-aware reverse attention.
%
%
Although this group excels in coarse-grained segmentation such as evident polyps or skin lesions, it struggles with precise segmentation in scenarios characterized by high-resolution or complex pathological slices.

\IEEEpubidadjcol

Overall, we can succinctly infer that the aforementioned two groups primarily concentrate on different tiers of features in deep supervised learning. Nevertheless, both high-level semantic and low-level pathological features are indispensable in the examination of medical images. In response to this issue, we propose an innovative learning paradigm that simultaneously captures coarse-grained semantic attributes and fine-grained detail features, as illustrated in Figure\,\ref{fig:motivation}(c).  Specifically, we introduce the Detail Enhancement Module (DEM) and the Semantic Enhancement Module (SEM), which operate concurrently during the decoding stage. DEM and SEM respectively apply detail and semantic masks derived from low-level and high-level features to enhance the feature supervision during decoding.
%
%
Such that, a synchronized amalgamation of both low-level (texture, color, and contrast) and high-level (semantic and object) feature supervision is ensured, thereby yielding more reliable segmentation outcomes.

\begin{figure}
    \centering
    \includegraphics[width=1\linewidth]{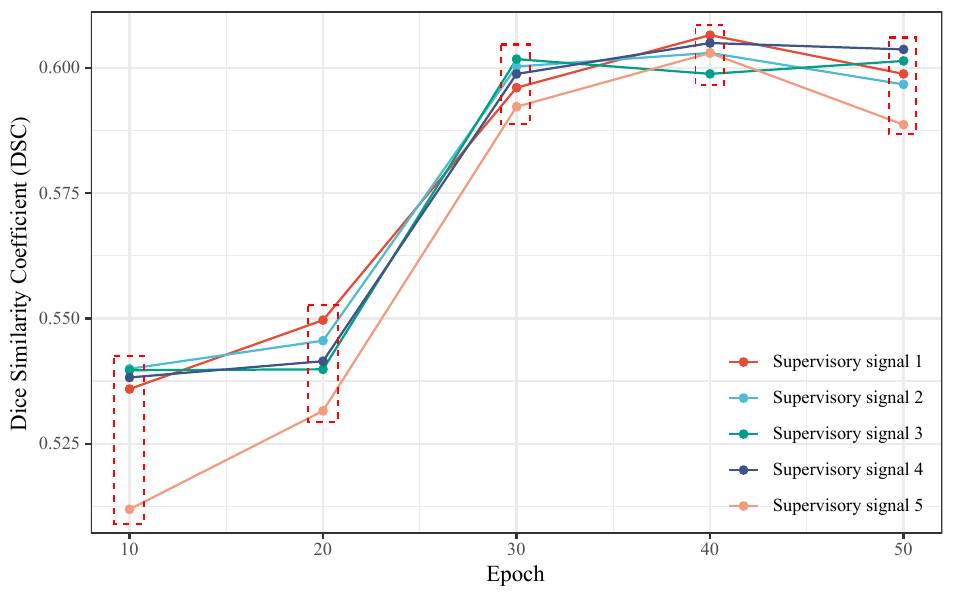}
    \caption{Quality rankings on CVC-ClonDB testset\cite{bernal2015wm} of supervision signals generated by UNet3+\cite{huang2020unet} over the first 50 training epochs. The index of the supervisory signals indicates their generation at distinct stages within the network, wherein a smaller index corresponds to a position closer to the input.
    The red dashed box underscores the non-heuristic nature of the quality concerning supervision signals at diverse stages.}
    \label{fig:motivation2}
    \vspace{-0.2cm}
\end{figure}

Beyond the mode of supervision, the allocation of supervisory loss magnitude at each stage also holds notable significance since the extraction and supervision of features at diverse scales inevitably exhibit varying degrees of complexity. 
Consequently, it becomes imperative to administer proper weights to the loss at distinct stages, with the aim of stabilizing the supervisory learning process and circumventing the propensity of descending into local optima.Existing works assign the weight of loss at each stage based on rule-of-thumb design. 
Particularly, semantic importance-based methods equally supervise all signals by assuming that they hold the same importance\cite{wang2023xbound}, whereas detail importance-based methods assign more penalties to output from stages proximal to the input layer\cite{ruan2023ege}. 
However, our observation in Figure \ref{fig:motivation2} shows that the quality of signals varies even from epochs.
Therefore, manually setting the loss weight for each stage necessitates substantial human expertise and may not invariably result in optimal supervisory effects. 
%
To mitigate this, we further introduce an innovative uncertainty-based adaptive loss, which assigns the weight of loss at each stage according to uncertainty estimation of the output features. Given that the uncertainty metric effectively mirrors the feature quality within clinical diagnosis contexts\cite{baumgartner2019phiseg,judge2022crisp}, our crafted loss strategy can autonomously bolster the supervision learning of each stage by adaptively distributing their loss weights.

\begin{figure*}
    \centering
    \includegraphics[width=1\linewidth]{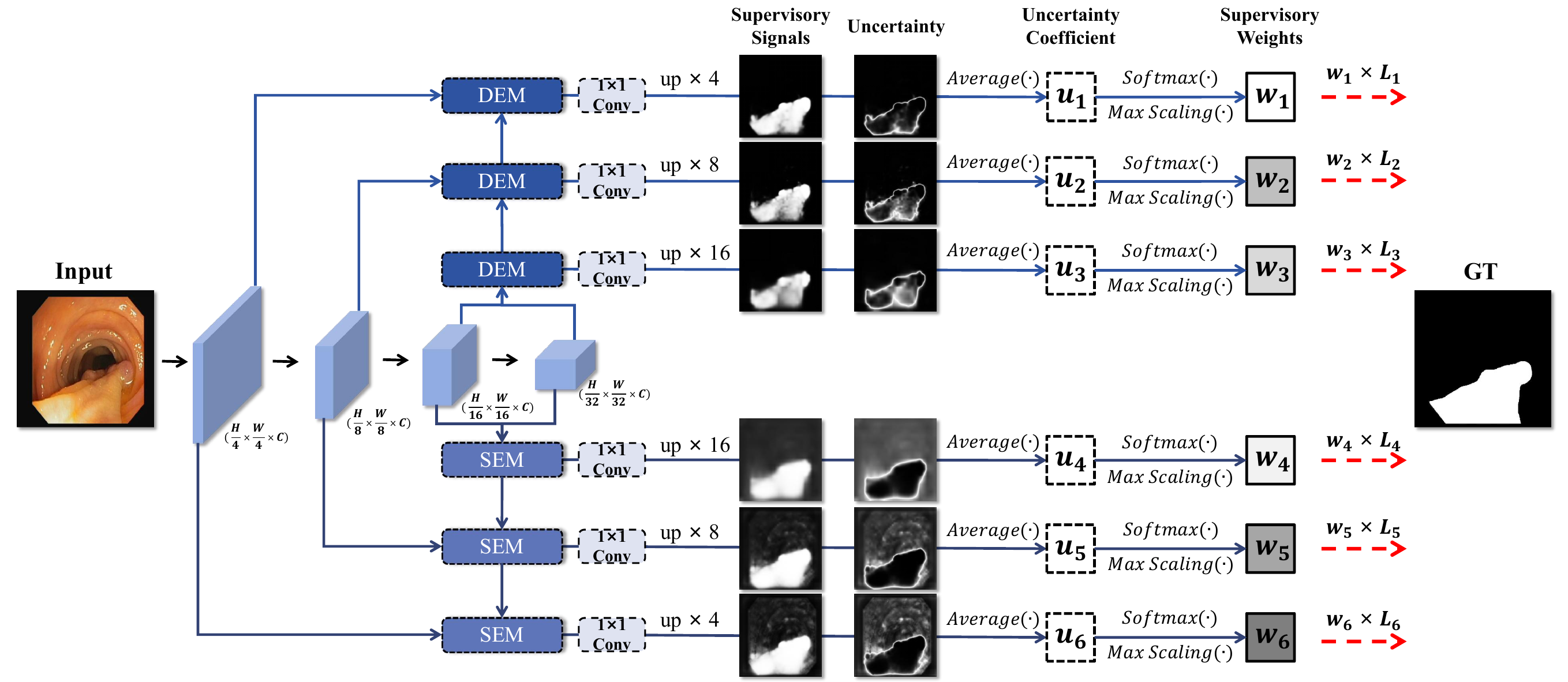}
    \caption{The framework of our proposed DS$^2$Net. We use the DEM and SEM to enhance intermediate features with distinct scales, where the outcomes are supervised with uncertainty-based adaptive loss. The red dash lines denote the deep supervision.}
    \label{fig:framework}
\end{figure*}

Our method, built upon 
DEM and SEM modules for medical image Segmentation, dubbed DS$^2$Net, is elucidated in Figure\,\ref{fig:framework}. DS$^2$Net adaptively learns high-level and low-level feature information at each stage under the supervision of uncertainty-based loss. Its efficacy is validated through extensive experiments across six benchmarks 
and results hightlight significantly improvement over existing methods\cite{rahman2023medical,tang2022duat,dong2021polyp,lou2022caranet,fan2020pranet,kim2021uacanet,ruan2023ege,zhou2018unet++,huang2020unet}.
Moreover, our experiments show the proposed adaptive supervision loss can also consistently heighten the performance of other existing deep supervision models, substantiating the efficacy of adaptive loss in deep supervision learning.Our contributions include:
%
%
\begin{itemize}
\item We propose DS$^2$Net, a novel deep supervision network that navigates both low-level detailed and high-level semantic features for medical image segmentation. 
\item We put forth an uncertainty-based supervision loss, the first non-heuristic loss to our best knowledge,  to adaptively assign the strength of supervision at each stage. 
\item Comparative experiments juxtaposed with leading-edge medical image segmentation models demonstrate the superior efficacy of our method across six datasets.
\end{itemize}

\section{Related Work}

\subsection{Medical Image Segmentation}
Driven by the triumph of FCN \cite{long2015fully} in semantic segmentation, the encoder-decoder-based architecture is liberally adopted within the medical imaging community.
For a more refined segmentation, UNet\cite{ronneberger2015u} enhances the fully connected decoder to an expansive one. Researchers subsequently amplifies the performance by incorporating dense connections\cite{li2018h}, residual optimization\cite{valanarasu2021kiu}, attention mechanisms\cite{wang2020non}, \emph{etc}, into the U-shaped architecture. The recent breakthrough of vision transformer (ViT)\cite{dosovitskiy2020image} spurs a new paradigm for medical image segmentation\cite{chen2021transunet,zhang2021transfuse}. Particularly, Swin-UNet\cite{cao2022swin} launches the first Transformer-based U-shaped network, albeit the massive computational costs. In response, EGE-UNet\cite{ruan2023ege} and MALUNet\cite{ruan2022malunet} further proffer lightweight segmentation networks for edge devices. Recent studies have also delved into model compression approaches including feature grouping\cite{wang2022uctransnet} and filter pruning\cite{dinsdale2022stamp} to further augment the efficiency of segmentation models.

\subsection{Deep Supervised Segmentation} 

The concept of deep supervision networks was initially introduced for image classification \cite{lee2015deeply}, targeting the optimization of feature learning while providing transparency into the learning process of hidden layers. On the basis of the supervision mode, prevailing methods can be broadly bifurcated into two classifications: architecture importance-based and signal importance-based approaches.
The former group employs supervision on the decoder's outputs within each stage of the U-shaped architecture, thereby learning pixel-level fine-grained features. For instance, UNet++\cite{zhou2018unet++} appends auxiliary outputs at diverse scales to guide feature learning. DNA\cite{liu2021dna} utilizes nonlinear aggregation for loss calculation, while EAC\cite{reiss2021every} improves the performance through multi-label loss. Conversely, signal importance-based methods focus on learning high-level intermediate semantic features,\emph{w.r.t}, supervisory signals. This is achieved by substituting the traditional decoder with global mapping, where the output supervisory signals are supervised at disparate hierarchical echelons. For example, UACANet\cite{kim2021uacanet} uses uncertain regions of saliency maps to enhance the edge information of intermediary outputs. 
%
Yue\emph{et al.}\cite{yue2023attention} designed a attention-guided pyramid network to refine the context feature of each layer.
In this paper, we present a novel DS$^2$Net, which learns both low-level detailed and high-level semantic features for deep supervised medical image segmentation.

\section{The Proposed Method}
An overview of DS$^2$Net is shown in Figure\,\ref{fig:framework}. Upon inputting a medical image, we initially extract four levels of feature maps of various scales sequentially utilizing the Pyramid Vision Transformer (PVT) block\cite{wang2021pyramid}. These features are subsequently funneled into the DEM and SEM for decoding, with the resulting outcomes being upscaled to generate supervisory signals. During the training phase, we employ uncertainty-based adaptive loss to direct the optimization of each supervisory signal. Amid inference, all supervisory signals are cumulatively added to obtain the final segmentation outcomes. Detailed expositions of DEM, SEM, and uncertainty-based adaptive supervision are presented in Sec.\ref{sec:dem}, Sec.\ref{sec:sem}, and Sec.\ref{sec:uas}, correspondingly. 

\begin{figure*}
    \centering
    \includegraphics[width=1\linewidth]{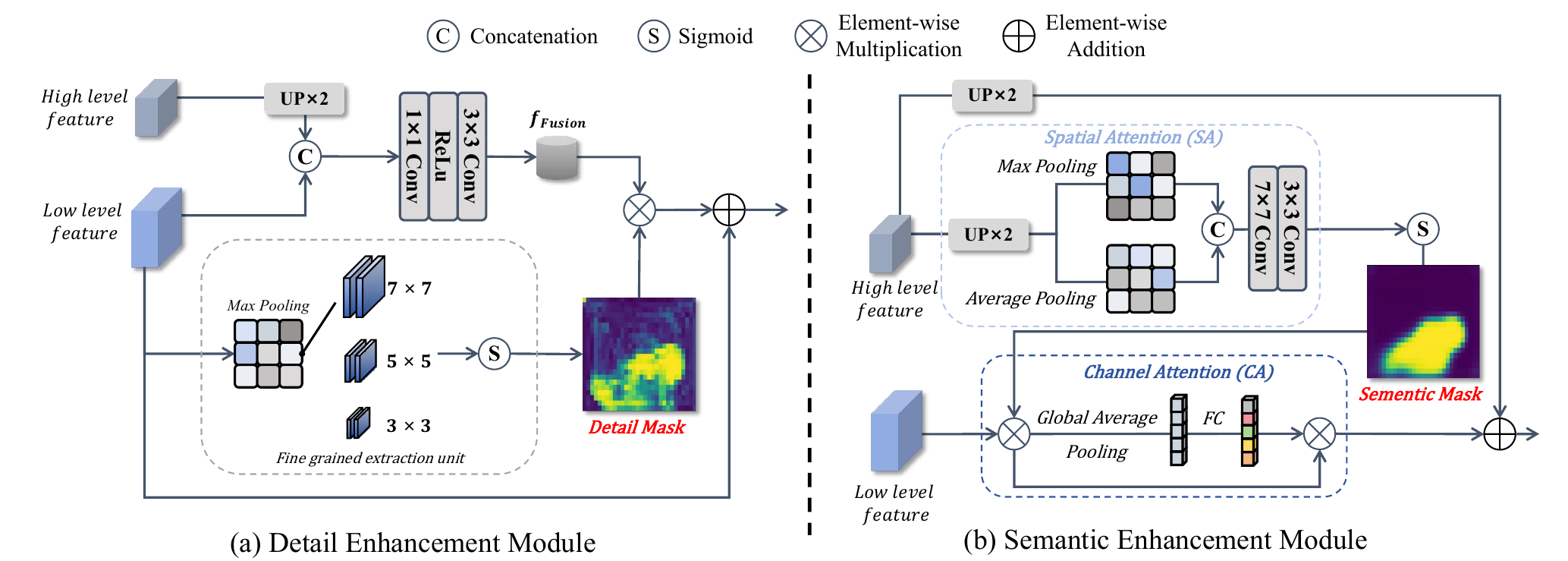}
    \caption{Overall architecture of Detail Enhancement Module (\textbf{\textit{Left}}) and Semantic Enhancement Module (\textbf{\textit{Right}}).}
    \label{fig:dem and sem}
\end{figure*}

\subsection{Detail Enhancement Module}\label{sec:dem}
Enhancing detailed supervision commonly requires the utilization of low-level features,\emph{w.r.t}, low-receptive-field feature maps located in the shallow layers of the encoding network. Low-level features are characterized by fine-grained information such as color, texture, and edges, thus providing valuable guidance for analyzing lesion boundaries and capturing minute pathological changes. However, existing methods simply rely on skip connections to fuse low-level and high-level features for supervision\cite{huang2020unet,qi2023mdf}, which fail to fully exploit the significant potential of low-level features. To mitigate this, we propose a Detail Enhancement Module (DEM), conceived to generate a novel detail mask aimed at further enhancing the fused features, as illustrated in Figure\,\ref{fig:dem and sem}(a). Particularly, DEM receives low-level feature $f_l$ and high-level feature $f_h$ from different depths of the encoding network as the inputs. The low-level feature $f_l$ is initially processed within a fine-grained extraction unit, composed of a maxpooling, two convolutions with the same kernel size, and  culminating with the application of a sigmoid function to garner a detail mask $M_d$ as: 
\begin{equation}
M_d=\sigma\left(Conv_{k}\left(Conv_{k}\left({Maxpool}\left(f_l\right)\right)\right)\right),
\end{equation}
where $\sigma\left(\cdot\right)$ and ${Maxpool}\left(\cdot\right)$ represent the sigmoid function and the maxpooling operation along the channel dimension, respectively, and $Conv_{k}$ delineates a convolutional layer with a kernel size of $k \times k$. The main objective of maxpooling is to preserve the most pertinent detail features while mitigating the impact of noise in low-level features. The double convolutions are designed to extend the receptive field. Given the various resolutions of low-level feature maps at disparate stages, we incorporate convolutional operations with distinct kernel sizes for each of the three DEM blocks in Figure\,\ref{fig:framework}. For example, the feature map possessing the utmost resolution employs a large 7 $\times$ 7 kernel.

The detail mask $M_d$ functions as a marker pinpointing fine-grained features within the entire image. Consecutively, we propose harnessing $M_d$ to augment the detail features by integrating it with complementary features fused from both high-level and low-level features. More explicitly, we first implement a 2$\times$ upsampling on the high-level feature $f_h$, succeeded by a concatenation operation alongside the low-level feature $f_l$. Next, we undertake a series of cascaded convolutional layers to further extract features. The initial layer deploys a 1 $\times$ 1 convolution with the intent of reducing the channel count, consecutively effectuated by a ReLU function. Furthermore, a 3 $\times$ 3 convolutional layer is deployed to accomplish fine-grained feature fusion, leading to the resultant fused feature $f_{Fusion}$ as:
\begin{equation}
f_{Fusion}=Conv_{3}\left(ReLU\left(Conv_{1}\left(\left[up\left(f_h\right), f_l\right]\right)\right)\right),
\end{equation}
where $\left[\cdot,\cdot\right]$ denotes the channel-wise concatenation operation, and $up\left(\cdot\right)$ represents the 2 $\times$ upsampling.
Ultimately, we conduct an element-wise multiplication between $M_d$ and $f_{Fusion}$, effectively amplifying the detailed information of the fused features. Consecutively, we assemble a residual connection encompassing the low-level features $f_l$ to generate the detail-enhanced decoder output $S_{d}$ as:
\begin{equation}
S_d =
M_d \odot f_{Fusion} + f_l.
\end{equation}

\subsection{Semantic Enhancement Module}\label{sec:sem}
Unlike the low-level counterparts, high-level features encapsulate substantial semantic information, comprising categories, context, and relationships. Consequently, it is pragmatic to supervise these high-level features for semantic enhancement, assisting in the precise localization of pathological objects amidst clinical diagnostics. Nevertheless, existing methods\cite{lou2022caranet,fan2020pranet,kim2021uacanet,dong2021polyp} simply adopt global mapping on high-level features to generate supervisory signals, which, to some degree, fails to thoroughly excavate the semantic information embedded within these high-level features.
To further augment this, we architect a Semantic Enhancement Module (SEM), incorporating a novel semantic mask derived from high-level features to heighten semantic supervision, as is illustrated in Figure\,\ref{fig:dem and sem}(b). Specifically, the semantic mask $M_s$ is first derived through the Spatial Attention (SA)\cite{woo2018cbam} block as 
\begin{equation}
M_s={SA}\left(up\left(f_h\right)\right),
\end{equation}
wherein the high-level feature $f_h$ is utilized as an input.
Thus, the semantic mask is multiplied with the low-level features $f_l$, culminating in semantically enhanced features. Given that high level features incorporate extensive contextual information intricately linked with channels\cite{zhao2019m2det}, we use Channel Attention (CA)\cite{wang2020eca} to encapsulate the associations among channels. Finally, we assemble a residual connection encompassing the high-level features $f_h$ to generate the final output of semantic-enhanced decoder  $S_{s}$:
\begin{equation}
S_s={CA}\left(M_s\odot f_l\right)+up\left(f_h\right).
\end{equation}

\subsection{Uncertainty-based Adaptive Supervision}\label{sec:uas}
In deep supervision learning, various stages of the network can spawn multiple supervisory signals, as depicted in Figure\,\ref{fig:framework}. Nevertheless, feature supervision at different stages inevitably exhibits a myriad of complexities, attributable to their diverse receptive scales and encapsulated information. Consequently, it becomes very crucial to assign suitable weights to the loss at distinct stages, with the objective of maintaining the stability of the supervisory learning process. Existing studies assign the loss weight at each stage based on rule-of-thumb design. For instance, Ruan\emph{et al.}\cite{ruan2023ege} proposed to confer more loss penalties to output from stages adjacent to the input layer, based on the assumption that shallow features are more challenging to learn. However, the heuristic design of loss weights demands substantial human expertise, and inevitably falls short of achieving optimal effects. To explain, the learning difficulty of different supervisory signals varies not only between stages, but also with the iterations in network training, as we quantitatively shown in Figure \ref{fig:motivation2}. 

\begin{table*}[t]
\setlength{\tabcolsep}{1pt}
\caption{Quantitative comparison of different methods on Kvasir-SEG (Kvasir), CVC-ClinicDB (ClinicDB), CVC-ColonDB (ColonDB) and ETIS-LaribPolypDB (ETIS) to validate our model's learning ability. $\uparrow$ denotes higher the better and $\downarrow$ denotes lower the better. Red indicates the best results and blue represents the second-best results. DS denotes the utilization of deep supervision.}
\centering
\resizebox{\linewidth}{!}{
\begin{tabular}{c|c|ccc|ccc|ccc|ccc}\hline
\multicolumn{1}{c|}{\multirow{2}{*}{Method}}& \multicolumn{1}{c|}{\multirow{2}{*}{DS}} & \multicolumn{3}{c|}{Kvasir (seen)} & \multicolumn{3}{c|}{ClinicDB (seen)} & \multicolumn{3}{c|}{ColonDB (unseen)} & \multicolumn{3}{c}{ETIS (unseen)}\\ 
\cline{3-14} && mDice$\uparrow$ & mIoU$\uparrow$ & $MAE$$\downarrow$ & mDice$\uparrow$ & mIoU$\uparrow$ & $MAE$$\downarrow$  & mDice$\uparrow$ & mIoU$\uparrow$ & $MAE$$\downarrow$ & mDice$\uparrow$ & mIoU$\uparrow$ & $MAE$$\downarrow$ \\ \hline
U-Net\cite{ronneberger2015u}&\scalebox{0.8}{-}&0.8566&0.7758&0.0404&0.8965&0.8359&0.0152&0.6134&0.5255&0.0480&0.4931&0.4059&0.0286\\
DoubleU-Net\cite{jha2020doubleu}&\scalebox{0.8}{-}&0.8927&0.8331&0.0273&0.9150&0.8604&0.0118&0.6967&0.6182&0.0416&0.6395&0.5641&0.0171\\
SANet\cite{wei2021shallow}&\scalebox{0.8}{-}&0.9041&0.8469&0.0283&0.9157&0.8593&0.0116&0.7521&0.6694&0.0432&0.7503&0.6540&0.0154\\
UCTransNet\cite{wang2022uctransnet}&\scalebox{0.8}{-}&0.8606&0.7831&0.0378&0.8696&0.8032&0.0164&0.6363&0.5454&0.0463&0.4937&0.4136&0.0277\\
UNet++\cite{zhou2018unet++}&\scalebox{0.8}{\Checkmark}&0.8498&0.7799&0.0427&0.8695&0.8066&0.0162&0.6221&0.5425&0.0468&0.4938&0.4249&0.0256\\
UNet3+\cite{huang2020unet}&\scalebox{0.8}{\Checkmark}&0.8597&0.7898&0.0405&0.9077&0.8557&0.0126&0.6369&0.5590&0.0486&0.5725&0.4964&0.0239\\
EGENet\cite{ruan2023ege}&\scalebox{0.8}{\Checkmark}&0.7597&0.6603&0.0700&0.7591&0.6677&0.0301&0.5266&0.4236&0.0629&0.3560&0.2794&0.0463\\
ACSNet\cite{zhang2020adaptive}&\scalebox{0.8}{\Checkmark}&0.8983&0.8378&0.0317&0.8821&0.8263&0.0113&0.7164&0.6492&0.0393&0.5784&0.5092&0.0593\\
UACANet\cite{kim2021uacanet}&\scalebox{0.8}{\Checkmark}&0.9050&0.8523&0.0264&0.9163&0.8703&0.0077&0.7830&0.7040&0.0336&0.6938&0.6155&0.0228\\
PraNet\cite{fan2020pranet}&\scalebox{0.8}{\Checkmark}&0.8982&0.8405&0.0296&0.8990&0.8492&0.0093&0.7115&0.6402&0.0431&0.6284&0.5666&0.0311\\
CaraNet\cite{lou2022caranet}&\scalebox{0.8}{\Checkmark}&0.9156&0.8653&\color{blue}{0.0226}&0.9363&0.8872&0.0069&0.7730&0.6891&0.0421&0.7477&0.6722&0.0170\\
Polyp-PVT\cite{dong2021polyp}&\scalebox{0.8}{\Checkmark}&0.9174&0.8642&0.0228&0.9368&\color{blue}{0.8894}&\color{blue}{0.0064}&\color{blue}{0.8083}&0.7274&0.0311&\color{blue}{0.7868}&\color{blue}{0.7058}&\textbf{\color{red}{0.0130}}\\
DuAT\cite{tang2022duat}&\scalebox{0.8}{\Checkmark}&0.9110&0.8597&0.0264&\color{blue}{0.9370}&\textbf{\color{red}{0.8900}}&0.0066&0.8059&0.7247&0.0357&0.7824&0.7052&0.0172\\
PVT-CASCADE\cite{rahman2023medical}&\scalebox{0.8}{\Checkmark}&\color{blue}{0.9192}&\color{blue}{0.8688}&0.0234&0.9343&0.8864&\textbf{\color{red}{0.0063}}&0.8061&\color{blue}{0.7278}&\color{blue}{0.0287}&0.7652&0.6875&0.0218\\
\rowcolor{gray!20}
DS$^2$Net (ours) &\scalebox{0.8}{\Checkmark}&\textbf{\color{red}{0.9272}}&\textbf{\color{red}{0.8785}}&\textbf{\color{red}{0.0218}}&\textbf{\color{red}{0.9373}}&\textbf{\color{red}{0.8900}}&0.0067&\textbf{\color{red}{0.8135}}&\textbf{\color{red}{0.7294}}&\textbf{\color{red}{0.0275}}&\textbf{\color{red}{0.7940}}&\textbf{\color{red}{0.7138}}&\color{blue}{0.0165}\\
\hline 
\end{tabular}}
\label{table: polyp}
\end{table*}

\begin{figure}[t]
    \centering
    \includegraphics[width=1\linewidth]{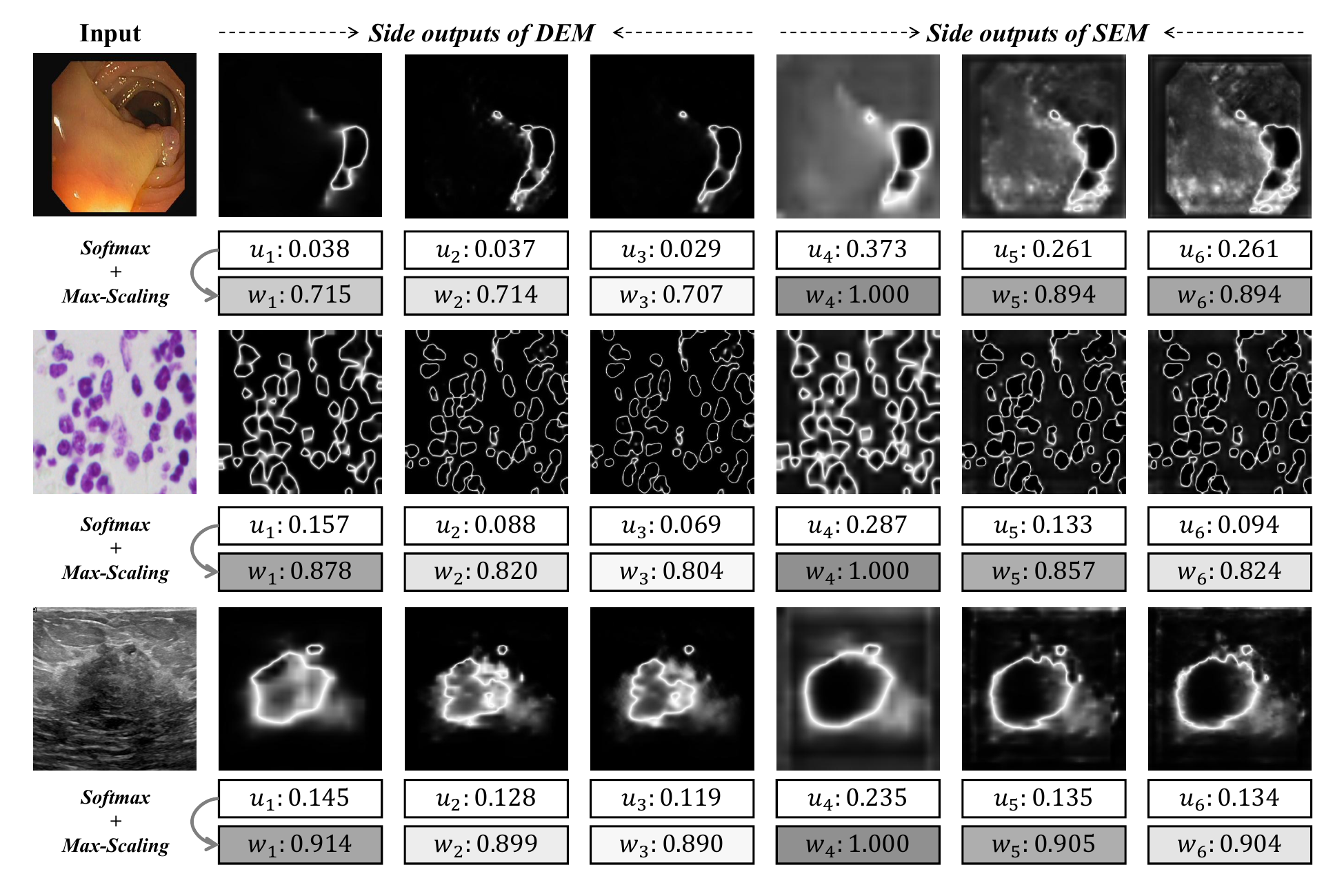}
    \caption{The uncertainty scores and supervision weights of different side outputs of DS$^2$Net at the first epoch.}
    \label{fig:uncertainty score}
\end{figure}

To address this conundrum, we introduce an uncertainty-based adaptive supervision method designed for an automatic allocation of loss weights to distinct supervisory signals. The primary motivation lies in the fact that the uncertainty of pixels serves as a reliable metric to assess signal quality within the context of medical image analysis. Therefore, we propose utilizing the indeterminacy of the supervisory signal as a metric for assessing the distribution of corresponding loss weights.
In particular, the supervisory signal can be firstly acquired by appending a 1×1 convolutional layer and performing an upsampling operation to the decoder,\emph{i.e.}, DEM or SEM.
Referring to all supervisory signals collectively as $p={p_1, p_2,  \ldots,p_N}$ ($N=6$ in DS$^2$Net), the uncertainty of a specific signal $p_i$ can be computed following prior work\cite{kim2021uacanet, zhang2020adaptive}:
\begin{equation}
u_i = AVG\left(1-\frac{\left|p_i-0.5\right|}{0.5}\right),
\label{eq:uncertainty}
\end{equation}
where $AVG\left(\cdot\right)$ denotes the average function.
As depicted in Figure\,\ref{fig:uncertainty score}, the acquired uncertainty scores are generally negligible yet exhibit considerable variability. While this substantiates our perspective that the quality of distinct signals varies, the initial values of $u = \{u_1, u_2, \ldots, u_N\}$ appear to be evidently inappropriate for deployment as supervisory weights. Thus, we apply the softmax function to $u$ to avert extreme distributions of uncertainty scores, resulting in the generation of $\bar{u}=\{\bar{u}_1, \bar{u}_2, \ldots, \bar{u}_N\}$ as:
\begin{equation}
\bar{u}_i=\frac{e^{u_i}}{\sum_{i=1}^n e^{u_i}}.
\label{eq: softmax}
\end{equation}

Furthermore, to emphasize highly uncertain supervisory signals, we perform MAX-Scaling on $\bar{u}$ to obtain supervisory weights $\mathrm{\Lambda}=\{\lambda_1, \lambda_2, \ldots, \lambda_N\}$ as:
\begin{equation}
\lambda_i = \frac{\bar{u}_i}{Max\{\bar{u}_1, \ldots, \bar{u}_n\}}.
\label{eq: max-scaling}
\end{equation}

Consequently, the loss function of our uncertainty-based adaptive supervision is defined as:
\begin{equation}
\mathcal{L}_{total}=\sum_{i=1}^N \lambda_i \times L(p_i,G),
\end{equation}
\begin{equation}
\mathcal{L}(p_i,G)=\mathcal{L}_{I o U}^w(p_i,G)+\mathcal{L}_{B C E}^w(p_i,G),
\label{eq: Loss}
\end{equation}
where $G$ represents the ground truth, $\mathcal{L}_{I o U}^w(p_i,G)$ and $\mathcal{L}_{B C E}^w(p_i,G)$ respectively denotes the weighted IoU loss\cite{mattyus2017deeproadmapper} and weighted binary cross entropy (BCE)\cite{de2005tutorial} loss.

\section{Experiments}
\subsection {Experimental Settings}
\textbf{Datasets and Evaluation protocols.} We conduct experiments on representative benchmarks for medical image segmentation with diverse image modalities. For images of colonoscopy polyps, the Kvasir-SEG\cite{jha2020kvasir} and CVC-ClinicDB\cite{silva2014toward} datasets are utilized. Kvasir-SEG contains 1000 images of assorted sizes, partitioned into a training set of 800 images and a testing set of 100 images (seen data). The CVC-ClinicDB dataset consists of 612 images, sized $384 \times 288$, inclusive of 488 training images, 62 validation images, and 62 testing images (seen data). To ascertain the generalization capability of the polyp segmentation model, we exclusively employ the CVC-ColonDB\cite{bernal2015wm} (comprising 380 images sized at $574 \times 500$) and the ETIS-LaribPolypDB\cite{vazquez2017benchmark} (consisting of 196 images sized $1225 \times 966$ images) as testing data (unseen data). For ultrasound images, we select the BUSI dataset\cite{al2020dataset}, which is centered on breast boundary restoration\cite{chen2022aau} and embodied by 780 images of varied sizes. For microscopic images, we leverage the 2018 Data Science Bowl challenge (2018-DSB) dataset\cite{caicedo2019nucleus} incorporating 670 images for the localization of nuclei. For quantitative comparison, we report three widely-used metrics including the mean Dice coefficient (mDice), mean Intersection over Union (mIoU), and mean absolute error (MAE). mDice and mIoU focus on the internal consistency of objects, while MAE represents the average value of the absolute error between the prediction and ground truth.

\textbf{Implementation Details.} We utilize a pre-trained PVT\cite{wang2021pyramid} model on ImageNet\cite{deng2009imagenet} as the backbone and conduct end-to-end training employing the AdamW optimizer\cite{loshchilov2017decoupled}. The initial learning rate is set to 1e-4 for datasets encompassing ultrasound and microscopic images, and 1e-5 for colonoscopy images. The batch size and weight decay are consistently set to 8 and 1e-4, respectively. For all training dataset, each image is resized to $352 \times 352$ without any data augmentation. Given the diverse scales of objects in medical imaging, a multi-scale training is adopted following previous work\cite{fan2020pranet,dong2021polyp,lou2022caranet,rahman2023medical}. Specifically, a scaling process is introduced for images within each batch. The scaling ratios for colonoscopy images are specified as 0.5, 1 and 1.5, whereas for ultrasound and microscopy images, the chosen ratios are 0.75, 1 and 1.25.
We give 150 epochs for training with colonoscopy images and 100 epochs with both ultrasound and microscopic images. To ensure a fair comparison, the evaluation results of the comparative models within this paper are derived using the officially provided open-source code or model. all experiments are carried out utilizing PyTorch\cite{paszke2019pytorch} on a singular NVIDIA GeForce RTX 4070 GPU boasting 12 GB of memory.

\begin{figure*}[htbp]
\centering
\includegraphics[width=1\textwidth]{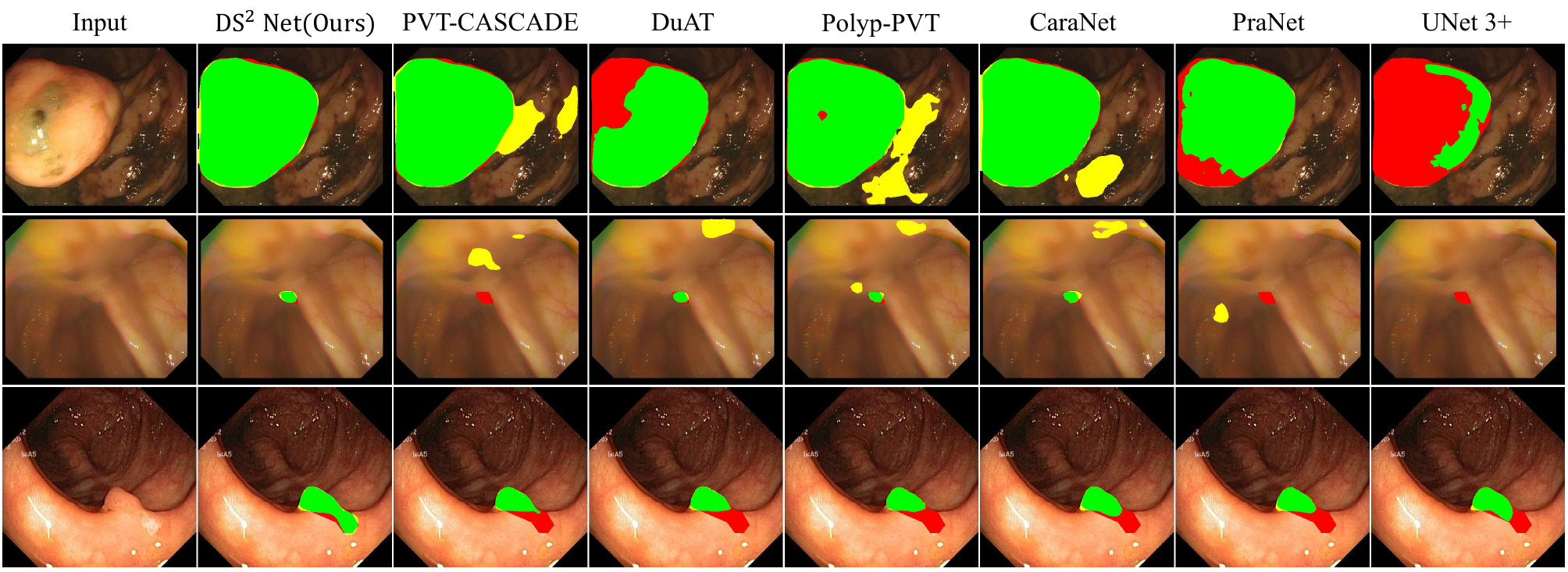} 
\caption{We select three representative polyp cases for qualitative analysis, which include: large polyp, small polyp and tissue-resembling polyp. Green indicates a correct polyp. Red is the missed polyp. Yellow is the redundant prediction.}
\label{fig:polyp qualitative analysis}
\end{figure*}

\subsection {Comparison with State-of-the-Art Methods}

\textbf{Results on Colonoscopy Images.} Table \ref{table: polyp} list the performance of DS$^2$Net for segment colonoscopy images in comparison with top-ranking deep supervision segmentation methods, including detail importance\cite{zhou2018unet++,huang2020unet,ruan2023ege} and semantic importance\cite{kim2021uacanet,fan2020pranet,lou2022caranet,dong2021polyp,tang2022duat,rahman2023medical} methods. 
We categorize test images that share the same source as the training data as seen data, and those from a different source as unseen data. On the seen data, our method yields the best performance among other methods. Especially, DS$^2$Net achieves 92.72\% mDice on the variable-sized Kvasir-SEG dataset\cite{jha2020kvasir}, furnishing a 0.8\% improvement over the contemporary state-of-the-art method, PVT-CASCADE\cite{rahman2023medical}.
On the unseen data, DS$^2$Net continues to upholds its superior performance. Notably, ETIS-LaribPolypDB\cite{vazquez2017benchmark} is the most challenging and high-resolution dataset among the other three. Nevertheless, DS$^2$Net emerges triumphant, delivering an impressive mDice score of 79.40\%. 
In Figure \ref{fig:polyp qualitative analysis}, we further show the qualitative results of different methods\cite{rahman2023medical,tang2022duat,dong2021polyp,lou2022caranet,fan2020pranet,huang2020unet}.
In the case of large polyp (first row), all methods are able to roughly segment the polyp, while our method exhibits the most similar results compared to the ground truth. In the case of small polyp and tissue-resembling polyp (second and third rows), DS$^2$Net is able to segment the object, even those that are difficult to distinguish by the naked eyes, while other methods exhibit instances of missing detection, false positives, or an inability to detect them.

\begin{table}[t]
\setlength{\tabcolsep}{2pt}
\caption{Quantitative comparison of different methods on BUSI~\cite{al2020dataset} and 2018-DSB~\cite{caicedo2019nucleus} to validate our model's learning ability. $\uparrow$ denotes higher the better and $\downarrow$ denotes lower the better. Red indicates the best results and blue represents the second-best results.}
\vspace{-0.2cm}
\centering
\resizebox{\linewidth}{!}{
\begin{tabular}{c|ccc|ccc}\hline
\multicolumn{1}{c|}{\multirow{2}{*}{Method}} & \multicolumn{3}{c|}{BUSI (Ultrasound)} & \multicolumn{3}{c}{2018-DSB (Microscope)}\\ 
\cline{2-7} & mDice$\uparrow$ & mIoU$\uparrow$ & $MAE$$\downarrow$ & mDice$\uparrow$ & mIoU$\uparrow$ & $MAE$$\downarrow$\\ \hline

U-Net\cite{ronneberger2015u} &0.6901&0.6034&0.0488&0.8555&0.7720&0.0342\\
UNet++\cite{zhou2018unet++} &0.6873&0.5935&0.0503&0.8806&0.8061&0.0322\\
UNet3+\cite{huang2020unet}&0.7055&0.6139&0.0493&0.8956&0.8214&0.0252\\
UACANet\cite{kim2021uacanet} &0.7473&0.6650&0.0442&0.8938&0.8141&0.0253\\
PraNet\cite{fan2020pranet} &0.7698&0.6847&0.0413&0.8897&0.8084&0.0273\\
DoubleU-Net\cite{jha2020doubleu} &0.7735&0.6870&0.0461&0.9011&0.8283&0.0255\\
SANet\cite{wei2021shallow} &0.7708&0.6842&0.0458&0.8905&0.8107&0.0267\\
UCTransNet\cite{wang2022uctransnet} &0.7681&0.6724&0.0551&0.8764&0.8043&0.0251\\
CaraNet\cite{lou2022caranet}&0.7769&0.6968&0.0383&0.8875&0.8053&0.0277\\
DuAT\cite{tang2022duat} &0.8017&0.7163&0.0406&\textcolor{blue}{0.9060}&\textcolor{blue}{0.8351}&0.0238\\
XBound-Former\cite{wang2023xbound} &0.7686&0.6843&0.0489&0.7555&0.6256&0.0605\\
PVT-CASCADE\cite{rahman2023medical} &\textcolor{blue}{0.8118}&\textcolor{blue}{0.7270}&\textcolor{blue}{0.0380}&0.9058&0.8337&\textcolor{blue}{0.0233}\\
\rowcolor{gray!20}
DS$^2$Net (ours) &\textbf{\textcolor{red}{0.8160}}&\textbf{\textcolor{red}{0.7286}}&\textbf{\textcolor{red}{0.0378}}&\textbf{\textcolor{red}{0.9097}}&\textbf{\textcolor{red}{0.8395}}&\textbf{\textcolor{red}{0.0226}}\\
\hline
\end{tabular}}
\label{table: busi}
\vspace{-0.3cm}
\end{table}

\begin{table*}
\setlength{\tabcolsep}{1pt}
\caption{Ablation study on DEM and SEM. }
\vspace{-0.2cm}
\centering
\resizebox{1\textwidth}{!}{
\begin{tabular}{c|cc|cc|cc|cc}
\hline
\multirow{2}{*}{Dataset} & \multicolumn{2}{c|}{\makecell[c]{PVT~\cite{wang2021pyramid}}} & \multicolumn{2}{c|}{\makecell[c]{PVT+DEM}}&\multicolumn{2}{c|}{\makecell[c]{PVT+SEM}}&\multicolumn{2}{c}{\makecell[c]{DS$^2$Net}}\\
\cline{2-9}

&mDice(\%)&mIoU(\%)&mDice(\%)&mIoU(\%)&mDice(\%)&mIoU(\%)&mDice(\%)&mIoU(\%) \\
\hline
Kvasir-SEG\cite{jha2020kvasir}&85.58&78.87&91.96&86.74&91.29&86.16&92.32\textbf{\textcolor{red}{(+6.74)}}&87.38\textbf{\textcolor{red}{(+8.51)}}\\
CVC-ClinicDB\cite{silva2014toward}&84.51&77.71&92.76&87.72&92.98&88.22&93.23\textbf{\textcolor{red}{(+8.72)}}&88.55\textbf{\textcolor{red}{(+10.84)}}\\
CVC-ColonDB\cite{bernal2015wm}&70.71&61.82&81.15&72.30&80.08&71.71&81.22\textbf{\textcolor{red}{(+10.51)}}&72.34\textbf{\textcolor{red}{(+10.52)}}\\
ETIS-LaribPolypDB\cite{vazquez2017benchmark}&67.49&58.19&76.05&67.99&78.53&70.78&79.02\textbf{\textcolor{red}{(+11.53)}}&71.06\textbf{\textcolor{red}{(+12.87)}}\\
BUSI\cite{al2020dataset}&76.58&67.57&81.00&72.19&81.26&72.23&81.40\textbf{\textcolor{red}{(+4.82)}}&72.72\textbf{\textcolor{red}{(+5.15)}}\\
2018-DSB\cite{caicedo2019nucleus}&88.01&80.22&90.60&83.52&90.50&83.31&90.70\textbf{\textcolor{red}{(+2.69)}}&83.65\textbf{\textcolor{red}{(+3.43)}}\\
\hline
\end{tabular}}
\label{table: ablation in dif component}
\vspace{-0.2cm}
\end{table*}

\begin{figure*}
    \centering
    \includegraphics[width=1\linewidth]{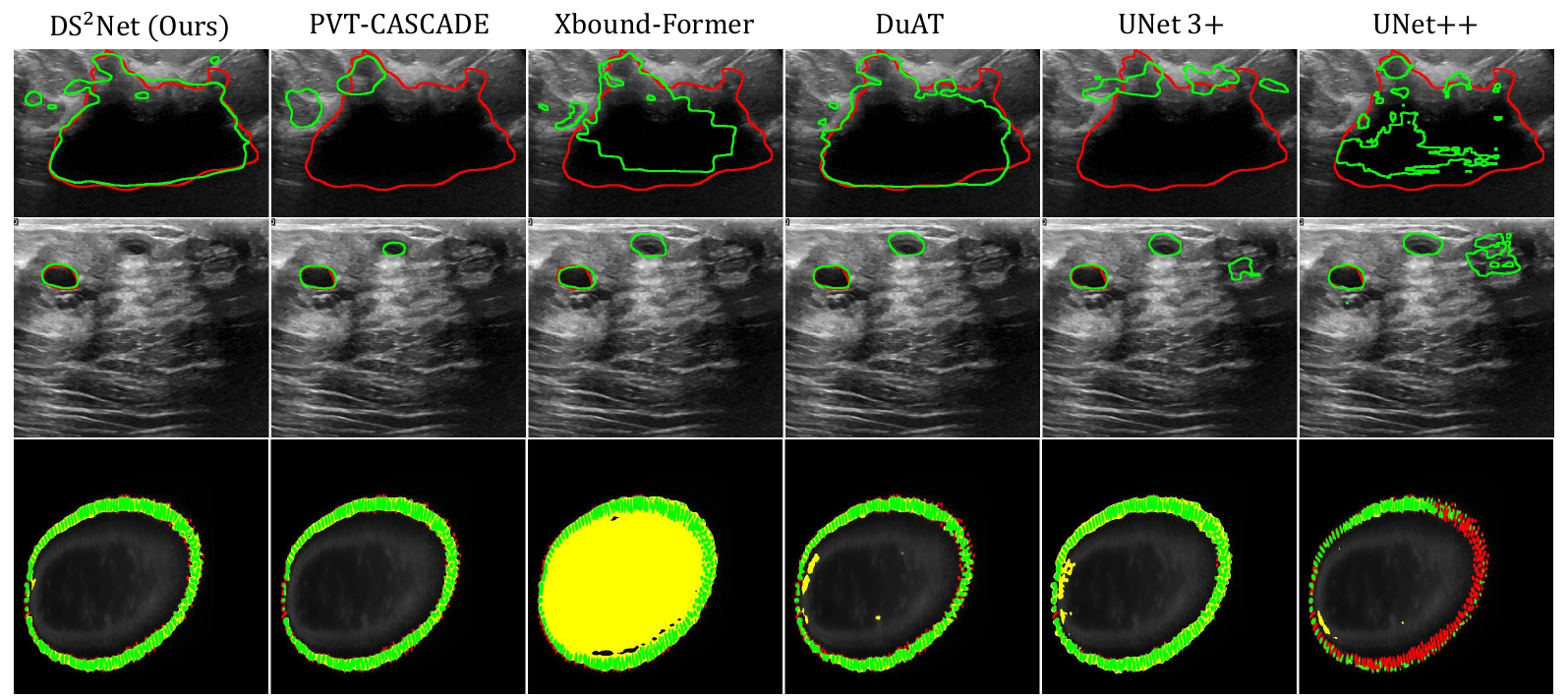}
    \caption{We select three representative cases from BUSI\cite{al2020dataset} and 2018-DSB\cite{caicedo2019nucleus}, which include: malignant tumor (BUSI), benign tumor (BUSI), and nucleus (2018-DSB). On BUSI dataset, the red curve is the ground-truth boundary. The green curve is the segmentation results of our method. On 2018-DSB dataset, green indicates a correct prediction, red is the missed prediction and yellow is the redundant prediction.}
    \label{fig:busi qualitative analysis}
\end{figure*}

\textbf{Results on Ultrasound Images.} The quantitative evaluation results on ultrasound images are presented in Table \ref{table: busi}. For breast lesions with indistinct boundaries, DS$^2$Net achieves consistent improvements against baseline models under comparable mDice, mIoU and MAE. The visualization of our method and comparative methods are shown in Figure \ref{fig:busi qualitative analysis}. Our method has the best performance in segmenting lesion. For instance, in the case of malignant tumor (first row, with serrated or lobulated boundary), other methods exhibit significant instances of false negatives, while our method addresses this issue well. These results verify the effectiveness of our DS$^2$Net in samples exhibiting blurred boundaries.

\textbf{Results on Microscopic Images.} In Table \ref{table: busi}, we also evaluate the performance of our method using the 2018-DSB dataset. Notably, we observe an enhancement of approximately 0.4\% in mDice and mIoU, while MAE decreased by about 0.1\%. In Figure \ref{fig:busi qualitative analysis}, we can see that our method achieves finer and more accurate predictions (more green, less red).

\textbf{Inference speed at different resolutions.} In Table~\ref{table:FPS}, we report the inference speed of DS$^2$Net and various types of deep supervision networks on datasets with varying resolutions. We note that our method strikes a balance between performance and inference speed. Specifically, the proposed model achieves higher FPS and mIoU scores compared to the semantic-based DSN~\cite{tang2022duat,rahman2023medical} with the same baseline, without significantly increasing memory cost. Furthermore, DS$^2$Net demonstrates more stable performance across different resolutions compared to detail-based DSN~\cite{huang2020unet,zhou2018unet++}.

\subsection {Ablation Study}

\textbf{Effectiveness of Different Network Components.} We use six datasets\cite{jha2020kvasir,caicedo2019nucleus,silva2014toward,bernal2015wm,vazquez2017benchmark,al2020dataset} to explore the core components in DS$^2$Net, as illustrated in Table \ref{table: ablation in dif component}. Notably, to fully showcase the performance of components, we have not employed adaptive supervision. Additionally, we report the performance of the network baseline, PVT~\cite{wang2021pyramid}.  We find that individual DEM or SEM can achieve an approximately 10\% improvement in mDice score on colonoscopy data, approaching the performance of state-of-the-art methods. Furthermore, we observe varying improvements from the two components on different datasets. For example, on the 2018-DSB dataset, DEM outperforms SEM by 0.21\% in mIoU, while SEM significantly outperforms DEM on ETIS-LaribPolypDB with a 2.48\% higher mDice score. This shows that discrete objects in microscope may benefit more from detail information, while high-resolution image require more semantic information.

\textbf{Influence of Different Operations in Adaptive Supervision.} We then investigate how different operations in adaptive supervision affects the performance on BUSI and 2018-DSB dataset. The results are reported in Table \ref{table:loss component}. We observe that directly using uncertainty scores as loss weights leads to a performance drop (72.72\%$\to$72.03\% mIoU and 83.65\%$\to$81.24\% mIoU). This indicates that high disparities in uncertainty scores in early training stage (see Figure \ref{fig:uncertainty score}) can easily trap the model in local optima. To address this issue, we introduce the softmax function to reduce the distances between uncertainty scores. However, we find that the performance remains poor performance on challenging BUSI dataset. The potential reason might be that the model converges too quickly (with small weight values), leading to the loss of important feature extraction. Therefore, we further employ Max-Scaling to amplify the weights, resulting in an improvement of approximately 0.2\% to 0.3\% in mDice and mIoU. The experiments show that adaptive supervisory loss can stably improve the results of our DS$^2$Net.

\begin{table}
\setlength{\tabcolsep}{1pt}
\caption{The FPS is tested on a NVIDIA GeForce RTX 4070 GPU on the CVC-ClinicDB (\textbf{\textit{low-resolution}}) and  ETIS-LaribPolypDB (\textbf{\textit{high-resolution}}), respectively.}
\centering
\resizebox{\linewidth}{!}{
\begin{tabular}{cccc|cc}
\hline
\multirow{2}{*}{Method} &\multicolumn{1}{c}{\multirow{2}{*}{Params(M) }}& \multicolumn{2}{c|}{\makecell[c]{Low-Resolution}}&\multicolumn{2}{c}{\makecell[c]{High-Resolution}}\\
\cline{3-6}

&&FPS &mIoU($\%$) &FPS &mIoU($\%$) \\
\hline
UNet++\cite{zhou2018unet++}&47.19&19.23&80.66&\textbf{23.63}&42.49\\
UNet3+~\cite{huang2020unet}&26.99&15.61&85.57&18.40&49.64\\
DUAT~\cite{tang2022duat}&24.95&32.80&\textbf{89.00}&22.55&70.52\\
PVT-CASCADE~\cite{rahman2023medical}&35.27&32.89&88.64&22.18&68.75\\
DS$^2$Net (ours)&24.91&\textbf{34.36}&\textbf{89.00}&23.22&\textbf{71.38}\\
\hline
\end{tabular}}
\label{table:FPS}
\end{table}

\textbf{Generality of Adaptive Supervision.} We further validate the generalization capability of our adaptive supervision by selecting three detail importance\cite{zhou2018unet++,huang2020unet,ruan2023ege} and semantic importance\cite{lou2022caranet,tang2022duat,rahman2023medical} deep supervision methods, respectively. As reported in Table \ref{table: play and plug}, adaptive supervision improves the performance of deep supervision methods. For instance, juxtaposed with the detail importance model UNet++, adaptive supervision boosts the mIoU score by 0.42\% (59.39\%$\to$59.77\%) and 1.05\% (80.61\%$\to$81.66\%) on the BUSI and 2018-DSB dataset. Similarly, the addition of adaptive supervision obtains much higher mDice and mIoU scores than the three semantic importance baselines. Through these empirical results, we demonstrate that the intervention of adaptive supervision remarkably elevates the performance of deep supervision networks.

\begin{table}[t]
\setlength{\tabcolsep}{1pt}
\caption{Ablation study on the adaptive supervision. \textit{$'$w/o Adaptive Supervision$'$} indicates that all supervision weights are equal (default to 1).}
\centering
\resizebox{\linewidth}{!}{
\begin{tabular}{c|cc|cc}
\hline
\multirow{2}{*}{Method} & \multicolumn{2}{c|}{\makecell[c]{BUSI}}&\multicolumn{2}{c}{\makecell[c]{2018-DSB}}\\
\cline{2-5}

&mDice($\%$)&mIoU($\%$)&mDice($\%$)&mIoU($\%$) \\
\hline
w/o Adaptive Supervision&81.40&72.72&90.70&83.65\\
\rowcolor{gray!20}
+Uncertainty&80.82\textcolor{blue}{(-0.58)}&72.03\textcolor{blue}{(-0.69)}&89.25\textcolor{blue}{(-1.45)}&81.24\textcolor{blue}{(-2.41)}\\
\rowcolor{gray!20}
+Softmax&80.87\textcolor{blue}{(-0.53)}&72.19\textcolor{blue}{(-0.53)}&90.85\textbf{\textcolor{red}{(+0.15)}}&83.74\textbf{\textcolor{red}{(+0.09)}}\\
\rowcolor{gray!20}
+Max-Scaling&81.60\textbf{\textcolor{red}{(+0.20)}}&72.86\textbf{\textcolor{red}{(+0.14)}}&90.97\textbf{\textcolor{red}{(+0.27)}}&83.95\textbf{\textcolor{red}{(+0.30)}}\\
\hline
\end{tabular}}
\label{table:loss component}
\end{table}

\textbf{Variants of DEM and SEM.} In Table \ref{table:ablation of DEM and SEM}, we present variations of DEM and SEM to elucidate the impact of high-level and low-level features across distinct modules. It can be observed that replacing the source of generating the attention mask in DEM with high-level features results in a performance drop. Similarly, the same phenomenon occurs in SEM. This indicates that low-level features effectively enhance detailed supervision, while high-level features contribute to the supervision of semantic attributes.

\section{Conclusion}
In this paper, we present a novel Detail-Semantic Deep Supervision Network (DS$^2$Net) for medical image segmentation. DS$^2$Net is built on a novel deep supervision paradigm capable of concurrently capturing coarse-grained semantic attributes and fine-grained detail features. This paradigm utilizes high-level and low-level features separately to generate semantic and detail masks, enhancing feature supervision. Additionally, diverging from the use of heuristic supervision loss prevalent in prior deep supervision method, we equip DS$^2$Net with an adaptive supervisory loss strategy that calculates uncertainty scores for various supervisory signals during training and accordingly directs the loss allocation based on these scores. Our experimental results substantiate profound improvement of DS$^2$Net in both quantitative and qualitative performance in comparison with state-of-the-arts. Equally important, the versatile adaptability of the adaptive supervisory loss strategy has also been well validated.

\begin{table}[t]
\setlength{\tabcolsep}{1pt}
\caption{The effect of adaptive supervision when applied to other deep supervision networks, including detail-based~\cite{zhou2018unet++,huang2020unet,ruan2023ege} and semantic-based~\cite{lou2022caranet,tang2022duat,rahman2023medical} methods.}
\centering
\resizebox{\linewidth}{!}{
\begin{tabular}{c|cc|cc}
\hline
\multirow{2}{*}{Method} & \multicolumn{2}{c|}{\makecell[c]{BUSI}}&\multicolumn{2}{c}{\makecell[c]{2018-DSB}}\\
\cline{2-5}

&mDice($\%$)&mIoU($\%$)&mDice($\%$)&mIoU($\%$) \\
\hline
UNet++\cite{zhou2018unet++}&68.73&59.35&88.06&80.61\\
\rowcolor{gray!20}
+Adaptive Supervision&68.84\textbf{\textcolor{red}{(+0.11)}}&59.77\textbf{\textcolor{red}{(+0.42)}}&89.23\textbf{\textcolor{red}{(+1.17)}}&81.66\textbf{\textcolor{red}{(+1.05)}}\\
UNet3+\cite{huang2020unet}&70.55&61.39&89.56&82.14\\
\rowcolor{gray!20}
+Adaptive Supervision&70.71\textbf{\textcolor{red}{(+0.16)}}&61.42\textbf{\textcolor{red}{(+0.03)}}&90.37\textbf{\textcolor{red}{(+0.81)}}&83.21\textbf{\textcolor{red}{(+1.07)}}\\
EGE-UNet\cite{ruan2023ege}&70.49&59.93&88.48&80.43\\
\rowcolor{gray!20}
+Adaptive Supervision&71.79\textbf{\textcolor{red}{(+1.30)}}&61.46\textbf{\textcolor{red}{(+1.53)}}&89.02\textbf{\textcolor{red}{(+0.54)}}&81.05\textbf{\textcolor{red}{(+0.62)}}\\
CaraNet\cite{lou2022caranet}&77.69&69.68&88.75&80.53\\
\rowcolor{gray!20}
+Adaptive Supervision&79.17\textbf{\textcolor{red}{(+1.48)}}&69.75\textbf{\textcolor{red}{(+0.07)}}&89.18\textbf{\textcolor{red}{(+0.43)}}&81.14\textbf{\textcolor{red}{(+0.61)}}\\
DuAT\cite{tang2022duat}&80.17&71.63&90.60&83.51\\
\rowcolor{gray!20}
+Adaptive Supervision&80.99\textbf{\textcolor{red}{(+0.82)}}&72.43\textbf{\textcolor{red}{(+0.80)}}&91.23\textbf{\textcolor{red}{(+0.63)}}&84.40\textbf{\textcolor{red}{(+0.89)}}\\
PVT-CASCADE\cite{rahman2023medical}&81.18&72.70&90.58&83.37\\
\rowcolor{gray!20}
+Adaptive Supervision&81.49\textbf{\textcolor{red}{(+0.31)}}&72.79\textbf{\textcolor{red}{(+0.09)}}&90.60\textbf{\textcolor{red}{(+0.02)}}&83.41\textbf{\textcolor{red}{(+0.04)}}\\
\hline
\end{tabular}}
\label{table: play and plug}
\vspace{-0.2cm}
\end{table}

\begin{table}[t]
\setlength{\tabcolsep}{1pt}
\caption{Quantitative comparison of variants of DEM and SEM.}
\centering
\resizebox{\linewidth}{!}{
\begin{tabular}{c|cc|cc}
\hline
\multirow{2}{*}{Method} & \multicolumn{2}{c|}{\makecell[c]{BUSI}}&\multicolumn{2}{c}{\makecell[c]{2018-DSB}}\\
\cline{2-5}
&mDice($\%$)&mIoU($\%$)&mDice($\%$)&mIoU($\%$) \\
\hline
DEM&81.00&72.19&90.60&83.52\\
\rowcolor{gray!20}
\makecell[c]{High-level Feature \\ $\to$ Attention Mask}&80.32\color{blue}{(-0.68)}&71.45\color{blue}{(-0.74)}&90.52\color{blue}{(-0.08)}&83.36\color{blue}{(-0.16)}\\
\hline
SEM&81.26&72.23&90.50&83.31\\
\rowcolor{gray!20}
\makecell[c]{Low-level Feature \\ $\to$ Attention Mask}&80.82\color{blue}{(-0.44)}&72.20\color{blue}{(-0.03)}&90.45\color{blue}{(-0.05)}&83.21\color{blue}{(-0.10)}\\
\hline
\end{tabular}}
\label{table:ablation of DEM and SEM}
\end{table}

\bibliography{main}

\begin{thebibliography}{10}
\providecommand{\url}[1]{#1}
\csname url@rmstyle\endcsname
\providecommand{\newblock}{\relax}
\providecommand{\bibinfo}[2]{#2}
\providecommand\BIBentrySTDinterwordspacing{\spaceskip=0pt\relax}
\providecommand\BIBentryALTinterwordstretchfactor{4}
\providecommand\BIBentryALTinterwordspacing{\spaceskip=\fontdimen2\font plus
\BIBentryALTinterwordstretchfactor\fontdimen3\font minus \fontdimen4\font\relax}
\providecommand\BIBforeignlanguage[2]{{%
\expandafter\ifx\csname l@#1\endcsname\relax
\typeout{** WARNING: IEEEtran.bst: No hyphenation pattern has been}%
\typeout{** loaded for the language `#1'. Using the pattern for}%
\typeout{** the default language instead.}%
\else
\language=\csname l@#1\endcsname
\fi
#2}}

\bibitem{al2020dataset}
W.~Al-Dhabyani, M.~Gomaa, H.~Khaled, and A.~Fahmy, ``Dataset of breast ultrasound images,'' \emph{Data in brief}, 2020.


\bibitem{baumgartner2019phiseg}
C.~F. Baumgartner, K.~C. Tezcan, K.~Chaitanya, A.~M. H{\"o}tker, U.~J. Muehlematter, K.~Schawkat, A.~S. Becker, O.~Donati, and E.~Konukoglu, ``Phiseg: Capturing uncertainty in medical image segmentation,'' in \emph{International Conference on Medical Image Computing and Computer-Assisted Intervention (MICCAI)}, 2019.


\bibitem{bernal2015wm}
J.~Bernal, F.~J. S{\'a}nchez, G.~Fern{\'a}ndez-Esparrach, D.~Gil, C.~Rodr{\'\i}guez, and F.~Vilari{\~n}o, ``Wm-dova maps for accurate polyp highlighting in colonoscopy: Validation vs. saliency maps from physicians,'' \emph{Computerized medical imaging and graphics}, 2015.


\bibitem{caicedo2019nucleus}
J.~C. Caicedo, A.~Goodman, K.~W. Karhohs, B.~A. Cimini, J.~Ackerman, M.~Haghighi, C.~Heng, T.~Becker, M.~Doan, C.~McQuin, \emph{et~al.}, ``Nucleus segmentation across imaging experiments: the 2018 data science bowl,'' \emph{Nature methods}, 2019.


\bibitem{cao2022swin}
H.~Cao, Y.~Wang, J.~Chen, D.~Jiang, X.~Zhang, Q.~Tian, and M.~Wang, ``Swin-unet: Unet-like pure transformer for medical image segmentation,'' in \emph{European Conference on Computer Vision (ECCV)}, 2022.


\bibitem{chen2022aau}
G.~Chen, L.~Li, Y.~Dai, J.~Zhang, and M.~H. Yap, ``Aau-net: an adaptive attention u-net for breast lesions segmentation in ultrasound images,'' \emph{IEEE Transactions on Medical Imaging (IEEE TMI)}, 2022.


\bibitem{chen2021transunet}
J.~Chen, Y.~Lu, Q.~Yu, X.~Luo, E.~Adeli, Y.~Wang, L.~Lu, A.~L. Yuille, and Y.~Zhou, ``Transunet: Transformers make strong encoders for medical image segmentation,'' \emph{arXiv preprint arXiv:2102.04306}, 2021.


\bibitem{de2005tutorial}
P.-T. De~Boer, D.~P. Kroese, S.~Mannor, and R.~Y. Rubinstein, ``A tutorial on the cross-entropy method,'' \emph{Annals of operations research}, 2005.


\bibitem{deng2009imagenet}
J.~Deng, W.~Dong, R.~Socher, L.-J. Li, K.~Li, and L.~Fei-Fei, ``Imagenet: A large-scale hierarchical image database,'' in \emph{IEEE/CVF Conference on Computer Vision and Pattern Recognition (CVPR)}, 2009.


\bibitem{dinsdale2022stamp}
N.~K. Dinsdale, M.~Jenkinson, and A.~I. Namburete, ``Stamp: Simultaneous training and model pruning for low data regimes in medical image segmentation,'' \emph{Medical Image Analysis}, 2022.


\bibitem{dong2021polyp}
B.~Dong, W.~Wang, D.-P. Fan, J.~Li, H.~Fu, and L.~Shao, ``Polyp-pvt: Polyp segmentation with pyramid vision transformers,'' \emph{arXiv preprint arXiv:2108.06932}, 2021.


\bibitem{dosovitskiy2020image}
A.~Dosovitskiy, L.~Beyer, A.~Kolesnikov, D.~Weissenborn, X.~Zhai, T.~Unterthiner, M.~Dehghani, M.~Minderer, G.~Heigold, S.~Gelly, \emph{et~al.}, ``An image is worth 16x16 words: Transformers for image recognition at scale,'' \emph{arXiv preprint arXiv:2010.11929}, 2020.


\bibitem{duan2022evaluating}
J.~Duan, M.~Bernard, L.~Downes, B.~Willows, X.~Feng, W.~F. Mourad, W.~St~Clair, and Q.~Chen, ``Evaluating the clinical acceptability of deep learning contours of prostate and organs-at-risk in an automated prostate treatment planning process,'' \emph{Medical Physics}, 2022.


\bibitem{fan2020pranet}
D.-P. Fan, G.-P. Ji, T.~Zhou, G.~Chen, H.~Fu, J.~Shen, and L.~Shao, ``Pranet: Parallel reverse attention network for polyp segmentation,'' in \emph{International Conference on Medical Image Computing and Computer-Assisted Intervention (MICCAI)}, 2020.


\bibitem{fu2018joint}
H.~Fu, J.~Cheng, Y.~Xu, D.~W.~K. Wong, J.~Liu, and X.~Cao, ``Joint optic disc and cup segmentation based on multi-label deep network and polar transformation,'' \emph{IEEE Transactions on Medical Imaging (IEEE TMI)}, 2018.


\bibitem{huang2020unet}
H.~Huang, L.~Lin, R.~Tong, H.~Hu, Q.~Zhang, Y.~Iwamoto, X.~Han, Y.-W. Chen, and J.~Wu, ``Unet 3+: A full-scale connected unet for medical image segmentation,'' in \emph{IEEE International Conference on Acoustics, Speech and Signal Processing (ICASSP)}, 2020.


\bibitem{jha2020doubleu}
D.~Jha, M.~A. Riegler, D.~Johansen, P.~Halvorsen, and H.~D. Johansen, ``Doubleu-net: A deep convolutional neural network for medical image segmentation,'' in \emph{IEEE International Symposium on Computer-Based Medical Systems}, 2020.


\bibitem{jha2020kvasir}
D.~Jha, P.~H. Smedsrud, M.~A. Riegler, P.~Halvorsen, T.~de~Lange, D.~Johansen, and H.~D. Johansen, ``Kvasir-seg: A segmented polyp dataset,'' in \emph{International Conference on Multimedia Modeling (MMM)}, 2020.


\bibitem{judge2022crisp}
T.~Judge, O.~Bernard, M.~Porumb, A.~Chartsias, A.~Beqiri, and P.-M. Jodoin, ``Crisp-reliable uncertainty estimation for medical image segmentation,'' in \emph{International Conference on Medical Image Computing and Computer-Assisted Intervention (MICCAI)}, 2022.


\bibitem{kim2021uacanet}
T.~Kim, H.~Lee, and D.~Kim, ``Uacanet: Uncertainty augmented context attention for polyp segmentation,'' in \emph{ACM International Conference on Multimedia (ACM MM)}, 2021.


\bibitem{lee2015deeply}
C.-Y. Lee, S.~Xie, P.~Gallagher, Z.~Zhang, and Z.~Tu, ``Deeply-supervised nets,'' in \emph{Artificial intelligence and statistics}, 2015.


\bibitem{li2018h}
X.~Li, H.~Chen, X.~Qi, Q.~Dou, C.-W. Fu, and P.-A. Heng, ``H-denseunet: hybrid densely connected unet for liver and tumor segmentation from ct volumes,'' \emph{IEEE Transactions on Medical Imaging (IEEE TMI)}, 2018.


\bibitem{lian2018hierarchical}
C.~Lian, M.~Liu, J.~Zhang, and D.~Shen, ``Hierarchical fully convolutional network for joint atrophy localization and alzheimer's disease diagnosis using structural mri,'' \emph{IEEE Transactions on Pattern Analysis and Machine Intelligence (IEEE TPAMI)}, 2018.


\bibitem{liu2021dna}
Y.~Liu, M.-M. Cheng, X.-Y. Zhang, G.-Y. Nie, and M.~Wang, ``Dna: Deeply supervised nonlinear aggregation for salient object detection,'' \emph{IEEE Transactions on Cybernetics}, 2021.


\bibitem{long2015fully}
J.~Long, E.~Shelhamer, and T.~Darrell, ``Fully convolutional networks for semantic segmentation,'' in \emph{IEEE/CVF Conference on Computer Vision and Pattern Recognition (CVPR)}, 2015.


\bibitem{loshchilov2017decoupled}
I.~Loshchilov and F.~Hutter, ``Decoupled weight decay regularization,'' \emph{arXiv preprint arXiv:1711.05101}, 2017.


\bibitem{lou2022caranet}
A.~Lou, S.~Guan, H.~Ko, and M.~H. Loew, ``Caranet: Context axial reverse attention network for segmentation of small medical objects,'' in \emph{Medical Imaging 2022: Image Processing}, 2022.


\bibitem{mattyus2017deeproadmapper}
G.~M{\'a}ttyus, W.~Luo, and R.~Urtasun, ``Deeproadmapper: Extracting road topology from aerial images,'' in \emph{International Conference on Computer Vision (ICCV)}, 2017.


\bibitem{mukherjee2020shallow}
P.~Mukherjee, M.~Zhou, E.~Lee, A.~Schicht, Y.~Balagurunathan, S.~Napel, R.~Gillies, S.~Wong, A.~Thieme, A.~Leung, \emph{et~al.}, ``A shallow convolutional neural network predicts prognosis of lung cancer patients in multi-institutional computed tomography image datasets,'' \emph{Nature machine intelligence}, 2020.


\bibitem{paszke2019pytorch}
A.~Paszke, S.~Gross, F.~Massa, A.~Lerer, J.~Bradbury, G.~Chanan, T.~Killeen, Z.~Lin, N.~Gimelshein, L.~Antiga, \emph{et~al.}, ``Pytorch: An imperative style, high-performance deep learning library,'' \emph{Conference on Neural Information Processing Systems (NeurIPS)}, 2019.


\bibitem{qi2023mdf}
W.~Qi, H.~Wu, and S.~Chan, ``Mdf-net: A multi-scale dynamic fusion network for breast tumor segmentation of ultrasound images,'' \emph{IEEE Transactions on Image Processing (IEEE TIP)}, 2023.


\bibitem{rahman2023medical}
M.~M. Rahman and R.~Marculescu, ``Medical image segmentation via cascaded attention decoding,'' in \emph{IEEE/CVF Winter Conference on Applications of Computer Vision (WACV)}, 2023.


\bibitem{reiss2021every}
S.~Reiss, C.~Seibold, A.~Freytag, E.~Rodner, and R.~Stiefelhagen, ``Every annotation counts: Multi-label deep supervision for medical image segmentation,'' in \emph{IEEE/CVF Conference on Computer Vision and Pattern Recognition (CVPR)}, 2021.


\bibitem{ronneberger2015u}
O.~Ronneberger, P.~Fischer, and T.~Brox, ``U-net: Convolutional networks for biomedical image segmentation,'' in \emph{International Conference on Medical Image Computing and Computer-Assisted Intervention (MICCAI)}, 2015.


\bibitem{ruan2022malunet}
J.~Ruan, S.~Xiang, M.~Xie, T.~Liu, and Y.~Fu, ``Malunet: A multi-attention and light-weight unet for skin lesion segmentation,'' in \emph{IEEE International Conference on Bioinformatics and Biomedicine (BIBM)}, 2022.


\bibitem{ruan2023ege}
J.~Ruan, M.~Xie, J.~Gao, T.~Liu, and Y.~Fu, ``Ege-unet: an efficient group enhanced unet for skin lesion segmentation,'' in \emph{International Conference on Medical Image Computing and Computer-Assisted Intervention (MICCAI)}, 2023.


\bibitem{silva2014toward}
J.~Silva, A.~Histace, O.~Romain, X.~Dray, and B.~Granado, ``Toward embedded detection of polyps in wce images for early diagnosis of colorectal cancer,'' \emph{International journal of computer assisted radiology and surgery}, 2014.


\bibitem{tang2022duat}
F.~Tang, Q.~Huang, J.~Wang, X.~Hou, J.~Su, and J.~Liu, ``Duat: Dual-aggregation transformer network for medical image segmentation,'' \emph{arXiv preprint arXiv:2212.11677}, 2022.


\bibitem{valanarasu2021kiu}
J.~M.~J. Valanarasu, V.~A. Sindagi, I.~Hacihaliloglu, and V.~M. Patel, ``Kiu-net: Overcomplete convolutional architectures for biomedical image and volumetric segmentation,'' \emph{IEEE Transactions on Medical Imaging (IEEE TMI)}, 2021.


\bibitem{vazquez2017benchmark}
D.~V{\'a}zquez, J.~Bernal, F.~J. S{\'a}nchez, G.~Fern{\'a}ndez-Esparrach, A.~M. L{\'o}pez, A.~Romero, M.~Drozdzal, A.~Courville, \emph{et~al.}, ``A benchmark for endoluminal scene segmentation of colonoscopy images,'' \emph{Journal of healthcare engineering}, 2017.


\bibitem{wang2022uctransnet}
H.~Wang, P.~Cao, J.~Wang, and O.~R. Zaiane, ``Uctransnet: rethinking the skip connections in u-net from a channel-wise perspective with transformer,'' in \emph{AAAI Conference on Artificial Intelligence (AAAI)}, 2022.


\bibitem{wang2023xbound}
J.~Wang, F.~Chen, Y.~Ma, L.~Wang, Z.~Fei, J.~Shuai, X.~Tang, Q.~Zhou, and J.~Qin, ``Xbound-former: Toward cross-scale boundary modeling in transformers,'' \emph{IEEE Transactions on Medical Imaging (IEEE TMI)}, 2023.


\bibitem{wang2020eca}
Q.~Wang, B.~Wu, P.~Zhu, P.~Li, W.~Zuo, and Q.~Hu, ``Eca-net: Efficient channel attention for deep convolutional neural networks,'' in \emph{IEEE/CVF Conference on Computer Vision and Pattern Recognition (CVPR)}, 2020.


\bibitem{wang2021pyramid}
W.~Wang, E.~Xie, X.~Li, D.-P. Fan, K.~Song, D.~Liang, T.~Lu, P.~Luo, and L.~Shao, ``Pyramid vision transformer: A versatile backbone for dense prediction without convolutions,'' in \emph{International Conference on Computer Vision (ICCV)}, 2021.


\bibitem{wang2020non}
Z.~Wang, N.~Zou, D.~Shen, and S.~Ji, ``Non-local u-nets for biomedical image segmentation,'' in \emph{AAAI Conference on Artificial Intelligence (AAAI)}, 2020.


\bibitem{wei2021shallow}
J.~Wei, Y.~Hu, R.~Zhang, Z.~Li, S.~K. Zhou, and S.~Cui, ``Shallow attention network for polyp segmentation,'' in \emph{International Conference on Medical Image Computing and Computer-Assisted Intervention (MICCAI)}, 2021.


\bibitem{woo2018cbam}
S.~Woo, J.~Park, J.-Y. Lee, and I.~S. Kweon, ``Cbam: Convolutional block attention module,'' in \emph{European Conference on Computer Vision (ECCV)}, 2018.


\bibitem{yue2023attention}
G.~Yue, S.~Li, R.~Cong, T.~Zhou, B.~Lei, and T.~Wang, ``Attention-guided pyramid context network for polyp segmentation in colonoscopy images,'' \emph{IEEE Transactions on Instrumentation and Measurement}, 2023.


\bibitem{zhang2020adaptive}
R.~Zhang, G.~Li, Z.~Li, S.~Cui, D.~Qian, and Y.~Yu, ``Adaptive context selection for polyp segmentation,'' in \emph{International Conference on Medical Image Computing and Computer-Assisted Intervention (MICCAI)}, 2020.


\bibitem{zhang2021transfuse}
Y.~Zhang, H.~Liu, and Q.~Hu, ``Transfuse: Fusing transformers and cnns for medical image segmentation,'' in \emph{International Conference on Medical Image Computing and Computer-Assisted Intervention (MICCAI)}, 2021.


\bibitem{zhao2019m2det}
Q.~Zhao, T.~Sheng, Y.~Wang, Z.~Tang, Y.~Chen, L.~Cai, and H.~Ling, ``M2det: A single-shot object detector based on multi-level feature pyramid network,'' in \emph{AAAI Conference on Artificial Intelligence (AAAI)}, 2019.


\bibitem{zhou2018unet++}
Z.~Zhou, M.~M. Rahman~Siddiquee, N.~Tajbakhsh, and J.~Liang, ``Unet++: A nested u-net architecture for medical image segmentation,'' in \emph{Deep Learning in Medical Image Analysis and Multimodal Learning for Clinical Decision Support}, 2018.


\end{thebibliography}
\bibliographystyle{IEEEtran}

\newpage
\vspace{11pt}

\vfill

\end{document}